\documentclass[sigconf]{acmart}

\usepackage{algorithm}
\usepackage{algorithmic}
\usepackage{multicol}
\usepackage{multirow}
\usepackage{subfigure}
\usepackage{tikz}
\usepackage{amsmath}
\usepackage{enumitem}
\usepackage{booktabs}
\usepackage{bbding}
\usepackage{graphicx}
\usepackage{stfloats}
\usepackage{amsfonts}
\usepackage{babel}
\usepackage{balance}
\usepackage{xcolor}
\AtBeginDocument{%
  \providecommand\BibTeX{{%
    \normalfont B\kern-0.5em{\scshape i\kern-0.25em b}\kern-0.8em\TeX}}}

\copyrightyear{2023}
\acmYear{2023}
\setcopyright{acmlicensed}\acmConference[SIGIR '23]{Proceedings of the 46th International ACM SIGIR Conference on Research and Development in Information Retrieval}{July 23--27, 2023}{Taipei, Taiwan} \acmBooktitle{Proceedings of the 46th International ACM SIGIR Conference on Research and Development in Information Retrieval (SIGIR '23), July 23--27, 2023, Taipei, Taiwan}
\acmPrice{15.00}
\acmDOI{10.1145/3539618.3592073} \acmISBN{978-1-4503-9408-6/23/07}

\begin{document}

\title{Towards Robust Knowledge Tracing Models via \mbox{\emph{k}}-Sparse Attention}

\author{Shuyan Huang}
\affiliation{%
  \institution{Think Academy International Education\\TAL Education Group}
  \city{Beijing}
  \country{China}
}
\email{huangshuyan@tal.com}

\author{Zitao Liu}\authornote{The corresponding author: Zitao Liu.}
\affiliation{%
  \institution{Guangdong Institute of Smart Education\\Jinan University}
  \city{Guangzhou}
  \country{China}
}
\email{liuzitao@jnu.edu.cn}

\author{Xiangyu Zhao}
\affiliation{%
  \institution{Applied Machine Learning Lab\\School of Data Science\\City University of Hong Kong}
  \city{Hong Kong}
  \country{China}
}
\email{xianzhao@cityu.edu.hk}

\author{Weiqi Luo}
\affiliation{%
  \institution{Guangdong Institute of Smart Education\\Jinan University}
  \city{Guangzhou}
  \country{China}
}
\email{lwq@jnu.edu.cn}

\author{Jian Weng}
\affiliation{%
  \institution{College of Cyber Security\\Jinan University}
  \city{Guangzhou}
  \country{China}
}
\email{cryptjweng@gmail.com}

\renewcommand{\shortauthors}{Shuyan Huang, Zitao Liu, Xiangyu Zhao, Weiqi Luo, \& Jian Weng}

\begin{abstract}
Knowledge tracing (KT) is the problem of predicting students' future performance based on their historical interaction sequences. With the advanced capability of capturing contextual long-term dependency, attention mechanism becomes one of the essential components in many deep learning based KT (DLKT) models. In spite of the impressive performance achieved by these attentional DLKT models, many of them are often vulnerable to run the risk of overfitting, especially on small-scale educational datasets. Therefore, in this paper, we propose \textsc{sparseKT}, a simple yet effective framework to improve the robustness and generalization of the attention based DLKT approaches. Specifically, we incorporate a k-selection module to only pick items with the highest attention scores. We propose two sparsification heuristics : (1) soft-thresholding sparse attention and (2) top-$K$ sparse attention. We show that our \textsc{sparseKT} is able to help attentional KT models get rid of irrelevant student interactions and have comparable predictive performance when compared to 11 state-of-the-art KT models on three publicly available real-world educational datasets. To encourage reproducible research, we make our data and code publicly available at \url{https://github.com/pykt-team/pykt-toolkit}\footnote{We merged our model to the \textsc{pyKT} benchmark at \url{https://pykt.org/}.}.
\end{abstract}



\maketitle

\section{Introduction}
\label{sec:intro}

\begin{figure}[!tbph]
\vspace{-0.3cm}{}
\centering
\includegraphics[width=\columnwidth]{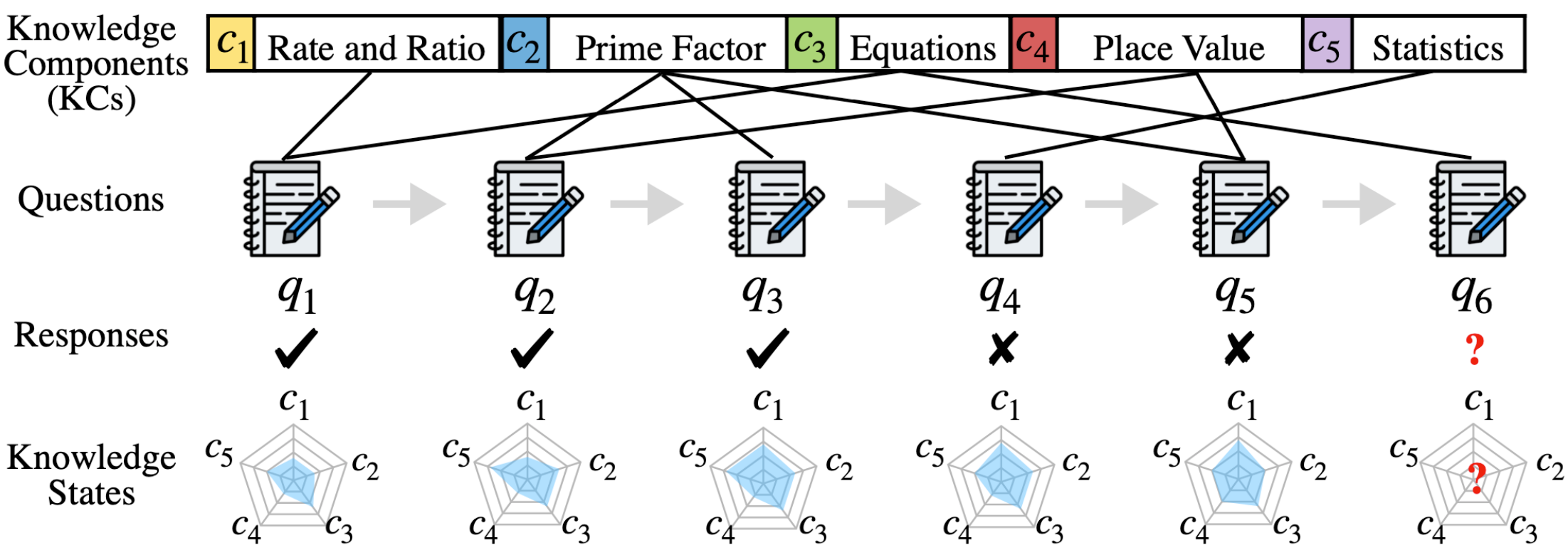}  \vspace{-0.6cm}
\caption{An illustration of the KT problem. A KC is a generality of everyday terms like concept, principle, or skill.}
\label{fig:kt_illustration}
\vspace{-0.4cm}
\end{figure}

The process of knowledge tracing involves utilizing a student's past learning interactions to construct a model to estimate his/her knowledge mastery to predict his/her future performance over a period of time (as shown in Figure \ref{fig:kt_illustration}). Such predictive abilities have the potential to improve students' learning outcomes and accelerate their progress when combined with high-quality learning materials and instructions. Recently, with the remarkable capability of attention mechanisms in natural language processing (NLP) or computer vision (CV) tasks, many DLKT models achieve accurate students' knowledge state estimations by utilizing attention netwo-rks to capture the intrinsic relevance among past interactions \cite{pandey2019self,ghosh2020context,liusimplekt}. 

Although the attentional DLKT approaches have achieved impressive results, they may run the risk of overfitting in real-world educational scenarios. Educational data is usually limited compared to large-scale language or image data. In the KT datasets, the question bank is usually bigger than the set of knowledge components (KCs) and a student has a very small number of question responses. Furthermore, since questions may be associated with limited relevant KCs, not all the past question responses contribute equally to the KT prediction task \cite{abdelrahman2019knowledge}. For instance, during predicting the student’s performance for $q_6$ in Figure \ref{fig:kt_illustration}, the KT models need easily look back to important historical information of the student's response in $q_1$ due to both of them are associated by $c_3$. Besides the question $q_1$, other questions likely have limited correlations to $q_6$. However, due to the smooth characteristic of the softmax function, irrelevant questions such as $q_1$ to $q_5$ may still get moderate attention scores in previous attention-based KT methods to predict the student's performance on $q_6$, which hinder the accurate inference of the student's knowledge state.

Therefore, in this work, to improve the robustness of attentional DLKT models meanwhile preserving the generalization performance under the assumption that it enables models to focus on the influential question inputs, we propose \textsc{sparseKT} that utilize sparse attention techniques to allow knowledge state estimations to be mapped to a limited number of pivotal interactions \cite{martins2016softmax, amini2022towards}. Specifically, our \textsc{sparseKT} approach focuses on the refinements of a popular attention based KT model: the Self Attentive Knowledge Tracing (SAKT) \cite{pandey2019self}. SAKT is a classical and widely used model for KT due to its relative simplicity, mathematically predictable behavior, and the fact that it handles the data sparsity problem based on relatively few past interactions. However, the pure attention mechanism in SAKT gives weights to each historical interaction of students, which may bring noise to the model and interfere with the accurate knowledge state estimations. Hence, we aim to develop a sparse attention framework with two simple but effective sparse attention heuristics to extract the relevant information from students' past learning sequences to perform better predictions when excluding interferences from other historical interactions. Our sparse attention methods refine the original dot-product attention by selecting $k$ most influential weights. We evaluate \textsc{sparseKT} on three benchmark datasets by comparing it with 11 previous approaches under a rigorous KT evaluation protocol \cite{liu2022pykt,chen2023improving,liu2023enhancing}. Experimental results demonstrate that our \textsc{sparseKT} approach achieves comparable prediction performance.

\section{Preliminary}
\label{sec:Preliminary}


\subsection{Self Attentive Knowledge Tracing}

 The self-attentive knowledge tracing (SAKT) model is the first method to use attention mechanisms in the context of KT \cite{pandey2019self}. The standard encoder in the Transformer model is employed in the basic setup of SAKT to extract context-aware interaction information through historical query and key-value pairs for the KT scenario \cite{vaswani2017attention}. The definitions of the query and key-value pairs are as follows:

\vspace{-0.3cm}
\begin{align}
& \mathbf{h}_{t+1} = \mbox{SelfAttention}(Q, K, V) \nonumber \\
& Q = \mathbf{z}_{t+1}; \quad K = \{\mathbf{y}_1, \cdots, \mathbf{y}_t\}; \quad V = \{\mathbf{y}_1, \cdots, \mathbf{y}_t\} \nonumber  \\ 
& \mathbf{y}_t = \mathbf{z}_t \oplus \mathbf{r}_t; \quad \mathbf{z}_t = \mathbf{W}^c \cdot \mathbf{e}^c_t; \quad \mathbf{r}_t = \mathbf{W}^q \cdot \mathbf{e}^q_t \label{eq:sakt}
\end{align}

\noindent where $\mathbf{h}_{t+1}$ is the learned context-aware latent representation that summarizes all available information for prediction at the $(t+1)$th timestamp. $\mathbf{z}_t$ and $\mathbf{y}_t$ denote the latent representations of the related KCs and interaction at timestamp $t$. $\mathbf{r}_t$ denotes the representation of student response. $\mathbf{e}^c_t$ is the \emph{n}-dimensional one-hot vector indicating the corresponding KC and $\mathbf{e}^q_t$ is the \emph{2}-dimensional one-hot vector indicating whether the question is answered correctly. $\mathbf{z}_t$ and $\mathbf{r}_t$ are \emph{d}-dimensional learnable real-valued vectors. $\mathbf{W}^c \in \mathbb{R}^{d \times n}$ and $\mathbf{W}^q \in \mathbb{R}^{d \times 2}$ are learnable linear transformation operations. $\oplus$ is the element-wise addition operator and $\cdot$ is the standard matrix/vector multiplication. 
$n$ is the total number of KCs. 

\subsection{Related Work}
\subsubsection{Attention based Knowledge Tracing}
Attention based KT models utilize attention mechanisms to capture the intrinsic dependencies among students’ chronologically ordered interactions. SAKT is the first research work that adopted a self-attention network to predict students' future performance \cite{pandey2019self}. Since then, many KT methods use attention networks to capture the potential relations between questions and responses \cite{choi2020towards,ghosh2020context,pandey2020rkt}. Choi et al. applied deep self-attentive layers in a pure Transformer architecture to extract the question and response representations \cite{choi2020towards}. Ghosh et al. proposed a monotonic attention mechanism that computes attention weights with exponential time-related decay \cite{ghosh2020context}.

\subsubsection{Sparse Attention}
Sparse attention improves the ordinary attention mechanism by only computing a limited selection of similarity scores from a sequence rather than all possible pairs \cite{martins2016softmax,child2019generating,zhao2019explicit,peters2019sparse}, which has shown promising performance in NLP and CV domains \cite{martins2016softmax,zhao2019explicit}. For example, Martins and Astudillo proposed a new activation function called sparsemax that is able to output sparse probabilities rather than traditional softmax \cite{martins2016softmax}. Child et al. introduced several sparse factorizations of the attention matrix without sacrificing performance \cite{child2019generating}. Zhao et al. designed an explicit sparse Transformer by selecting $k$ most relevant components \cite{zhao2019explicit}.


\section{The \textsc{sparseKT} Approach}
\label{sec:Method}

\begin{figure}[!thpb]
\vspace{-0.2cm}
    \includegraphics[width=\linewidth]{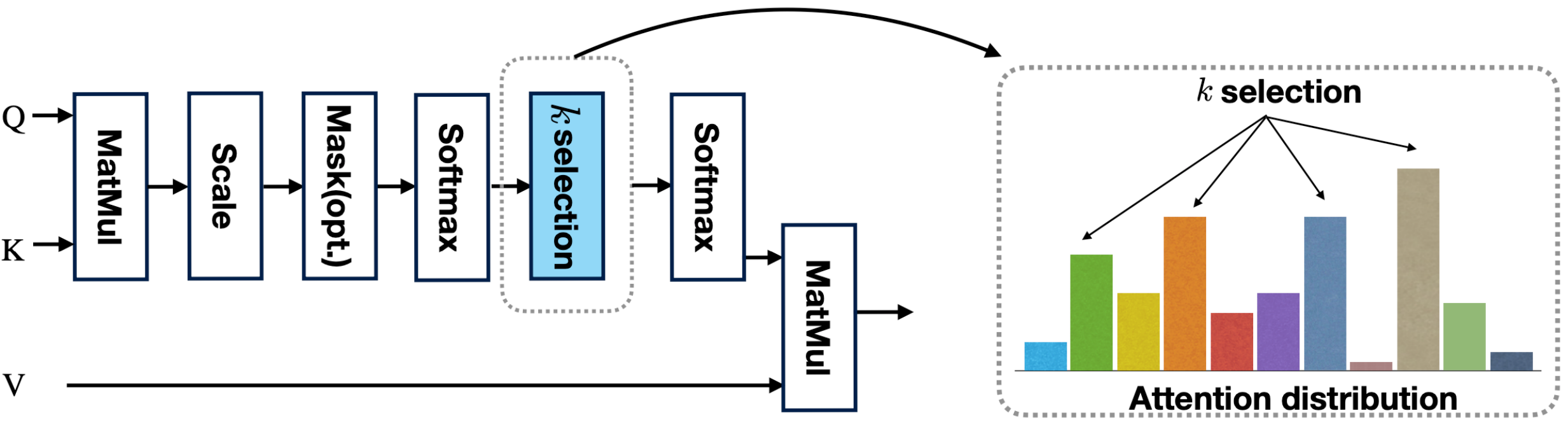}\vspace{-0.45cm}
    \caption{The \textsc{sparseKT} illustration.}
    \vspace{-0.5cm}
    \label{fig:framework}
\end{figure}

We propose that to improve the performance of attention-based knowledge tracing models, it is necessary to further enhance the generalization of the model. One way to accomplish this is by incorporating recent advancements in attention sparsification techniques. Therefore, in this paper, we propose \textsc{sparseKT}, a simple yet effective framework to facilitate the robustness of the attention based KT approach. Briefly, our \textsc{sparseKT} approach incorporates an additional $k$-sparse selection module after the standard self-attention function to only select the top $K$ interactions with highest attention scores. Only the selected $K$ interactions are used to make future predictions. The idea of \textsc{sparseKT} is illustrated in Figure \ref{fig:framework}.

\subsection{Embedding}
\label{sec:emb}
Inspired by the classic and simple Rasch model in psychometrics that explicitly uses a scalar to characterize the latent factor of question discrimination, we improve the SAKT's interaction representation (shown in eq.(\ref{eq:sakt})) by utilizing a question-specific discrimination factor to capture the individual differences among questions on the same KC. Specifically, let $\mathbf{x}_t$ be the enhanced representations that contain question-centric information, i.e.,

\vspace{-0.3cm}
$$\mathbf{x}_t = \mathbf{m}^q \odot \mathbf{v}^c \oplus \mathbf{z}_{t}$$

\noindent where $\mathbf{m}^q$ denotes the question-specific discrimination factor of question $q_t$ and $\mathbf{v}^c$ represents the variation of $q_t$ covering this KC set $\mathbf{c}$. Both $\mathbf{q}_t$ and $\mathbf{v}^c$ are \emph{d}-dimensional learnable real-valued vectors. $\odot$ is the element-wise product operator.

\subsection{\mbox{\emph{k}}-Sparse Attention}
\label{sec:sparse_attn}

In this section, we leverage sparsification techniques to enhance the generalization of KT models for better performance. In our work, historical interactions that have limited correlations to the current question will not be assigned to the attention scores. More specifically, we enhance the SAKT's scaled dot-product attention mechanism by using sparse attention to allow the model to focus on only a few pieces of historical information through explicit selection. Let $\mathcal{I} $ be the attention distribution computed from our question enhanced representations  $\mathbf{x}_{t+1}$, i.e., 

\vspace{-0.3cm}
\begin{equation*}
\mathcal{I} = \mbox{softmax}(\frac{QK^T}{\sqrt{d}}); Q = \mathbf{x}_{t+1}; K = \{\mathbf{x}_1, \cdots, \mathbf{x}_t\}
\end{equation*}

Let $\mathcal{M(\cdot)}$ be the mask operation that selects the $k$ most informative historical interactions, $\mathcal{I}_i$ denotes the attention score of $\mathbf{x}_i$. There are two implementations of $\mathcal{M(\cdot)}$ including soft-thresholding and top-$K$ sparse attention, described as follows:

\begin{itemize}[leftmargin=*]

\item	\textbf{soft-thresholding sparse attention}: we order all the attention scores $\mathcal{I}_{i}$ of attention distribution $\mathcal{I}$ from largest to smallest. And gradually pick up $\mathcal{I}_{i}$ into a weighting set $\mathcal{N}=\mathcal{N} \cup \{\mathcal{I}_{i}\}$ until the cumulative sum of $\mathcal{I}_{i}s$ in $\mathcal{N}$ is larger than the predefined soft-threshold $k$. Hence, we treat all the historical interactions with the attention scores $\mathcal{I}_{i}$ as the most contributive ones to predict a student's future performance. Other interactions are likely to be irrelevant to the prediction which will not be assigned attention scores by:
\begin{align*}
	\mathcal{M}_{\mbox{soft}}(\mathcal{I}_{i})=\left\{
		\begin{array}{l}
			\mathcal{I}_{i}  \qquad \text{ if } \mathcal{I}_{i} \in \mathcal{N} \\
		   -\infty   \qquad  \text{otherwise} \\
		\end{array}\right.
\end{align*}



\item	\textbf{top-$K$ sparse attention}: let $s$ be the $k$-th largest value in $\mathcal{I}$. We select the top $k$ largest scores of in $\mathcal{I}$ as the most influential components. Practically, if $\mathcal{I}_{i}$ is larger than $s$, we will select $\mathcal{I}_{i}$ and vice versa, i.e.,

\vspace{-0.3cm}
\begin{align*}
	\label{eq:topK}
	\mathcal{M}_{\mbox{topK}}(\mathcal{I}_{i})=\left\{
		\begin{array}{l}
			\mathcal{I}_{i} \qquad \text{ if }\;\mathcal{I}_{i}\ge s \\
		   -\infty \qquad \text{otherwise}  \\
		\end{array}\right.
\end{align*}


\end{itemize}

We then re-normalized the attention score distribution $I$, and the normalized scores of $\mathcal{I}_{i}$ in negative infinity are approximately 0. Therefore, we get a sparse attention distribution that explicitly chooses the highest attention scores that may influence the KT model decision. Finally, we obtain the knowledge state representation of $q_{t+1}$ as:


\vspace{-0.3cm}
\begin{equation*}
\mathbf{h}_{t+1} = \mbox{softmax}(\mathcal{M}(\mathcal{I}))V;  V = \{\mathbf{y}_1, \cdots, \mathbf{y}_t\} \nonumber\\
\end{equation*}

\vspace{-0.3cm}
\subsection{Prediction Layer}
\label{sec:pred}
We use a two-layer fully connected network to refine the knowledge state and the overall optimization function is as follows:


\begin{align*}
\hat{r}_{t+1} & = \sigma(\mathbf{w}^\top \cdot \mbox{ReLU} \bigl( \mathbf{W}_2 \cdot \mbox{ReLU} ( \mathbf{W}_1 \cdot [\mathbf{h}_{t+1}; \mathbf{x}_{t+1}] + \mathbf{b}_1 ) + \mathbf{b}_2 \bigl) + b) \\
\mathcal{L} & = - \sum_t \bigl( r_{t+1} \log \hat{r}_{t+1} \ + (1-r_t) \log (1-\hat{r}_{t+1}) \bigl) 
\end{align*}

\noindent where $\mathbf{W}_1$, $\mathbf{W}_2$, $\mathbf{w}$, $\mathbf{b}_1$, $\mathbf{b}_2$ and $b$ are trainable parameters and $\mathbf{W}_1 \in \mathbb{R}^{d \times 2d}$, 
$\mathbf{W}_2 \in \mathbb{R}^{d \times d}$, $\mathbf{w}, \mathbf{b}_1, \mathbf{b}_2 \in \mathbb{R}^{d \times 1}$, $b$ is scalar. $\sigma(\cdot)$ is the sigmoid function.

\section{Experiments}
\label{sec:exp}
We use three publicly educational datasets to evaluate the effectiveness of our model. We remove student sequences with fewer than 3 attempts and set the maximum length to 200 following the data preprocessing by \cite{liu2022pykt}. The datasets are described as follows:

\begin{itemize}[leftmargin=*]
	\item ASSISTments2015\footnote{\label{ft:l1}\quad\url{https://sites.google.com/site/assistmentsdata/datasets/ 2015-assistments-skill-builder-data.}} (AS2015): the dataset comprises mathematical exercises from the ASSISTments platform during the 2015 academic year. It ends up with 682,789 interactions, 19,292 sequences and 100 KCs after pre-processing.
	\item NeurIPS2020 Education Challenge\footnote{\quad\url{https://eedi.com/projects/neurips-education-challenge.}} (NIPS34): the dataset is provided by NeurlPS 2020 Education Challenge. We use the dataset of Task 3 \& Task 4 to evaluate our models \cite{wang2020instructions}. There are 1,399,470 interactions, 9,401 sequences, 948 questions, 57 KCs.
	\item Peking Online Judge\footnote{\quad\url{https://drive.google.com/drive/folders/1LRljqWfODwTYRMPw6wEJ_ mMt1KZ4xBDk.}} (POJ):  the dataset contains programming exercises on the Peking coding platform and is scraped by \cite{pandey2020rkt}. It has 987,593 interactions, 20,114 sequences and 2,748 questions.
\end{itemize}


We compare the two instances of our $k$-sparse self-attention framework, i.e., \textsc{sparseKT}-soft and \textsc{sparseKT}-topK to the following 11 KT models to evaluate the effectiveness of our approach:

\begin{itemize}[leftmargin=*]
\item \textit{DKT} \cite{piech2015deep}: uses an LSTM layer to encode the students' knowledge state for predicting their response performances.
\item \textit{DKT+} \cite{yeung2018addressing}: a variation of DKT that tackles the problems of reconstruction and non-consistent prediction.
\item \textit{KQN} \cite{lee2019knowledge}: calculates the relevance of student knowledge state encoder and skill encoder via the dot product.
\item \textit{DKVMN} \cite{zhang2017dynamic}: exploits two memory networks to extract the relations between different KCs and students' knowledge states.
\item \textit{ATKT} \cite{guo2021enhancing}: exploits adversarial perturbations to the interaction embeddings to enhance the models' generalization capability.
\item \textit{GKT} \cite{nakagawa2019graph}: utilizes graph neural networks to model the relation between KCs to predict the student's future performance on KCs.
\item \textit{SAKT} \cite{pandey2019self}: leverages a self-attention mechanism to capture relevance between exercises and responses.
\item \textit{SAINT} \cite{choi2020towards}: uses Transformer architecture to represent students' exercise and response sequences via encoder and decoder.
\item \textit{AKT} \cite{ghosh2020context}: introduces monotonic attention to enhance self-attention by considering the students' forgetting behavior.
\item \textit{HawkesKT} \cite{wang2021temporal}: utilizes the Hawkes process to model temporal cross-effects in student historical interactions.
\item \textit{IEKT} \cite{long2021tracing}: estimates students' knowledge states by modeling student cognition and knowledge acquisition behaviors.
\end{itemize}

\subsection{Results}
\label{sec:results}
\subsubsection{Overall Performance}

Table \ref{tab:overall} shows the overall performance of all models. From Table \ref{tab:overall}, we have the following observations: (1) compared to other baseline methods, our two \textsc{sparseKT} models almost always ranks top 3 in terms of AUC scores and accuracy on NIPS34 and POJ datasets. Although our \textsc{sparseKT} approaches are worse than several baselines such as DKT, DKT+, AKT on AS2015, the performance gaps are quite minimal and are mostly within a 0.7\% range. This indicates the strength of \textsc{sparseKT} as a baseline of KT; (2) in spite of our two \textsc{sparseKT} approaches are extensions of SAKT only by using sparse attention to replace scaled dot-product attention, they all have remarkable improvements of AUC scores by 0.97\%, 4.79\% and 1.38\% on three datasets on average compared to SAKT. This indicates our sparse attention mechanism allows KT models to pay attention to limited influential historical information that improves the predictive performance; and (3) comparing \textsc{sparseKT}-topK and \textsc{sparseKT}-soft, we can see, \textsc{sparseKT}-soft performs slightly better than \textsc{sparseKT}-topK. We believe that the student performance on questions still depends on several numbers of past interactions and the top-$K$ version of sparse attention may bring too limited information to hard to estimate a student's future performance compared to \textsc{sparseKT}-soft.
\begin{table}[!htbp] \vspace{-0.4cm}
\caption{AUC and accuracy results on AS2015, NIPS34 and POJ datasets. HAWKES and IEKT require both question IDs and KC IDs which are not available in AS2015 and POJ. }
\vspace{-0.3cm}
\tiny
\setlength\tabcolsep{1pt}
\begin{tabular}{lcccccc}
\toprule
\multicolumn{1}{c}{\multirow{2}{*}{Model}} & \multicolumn{3}{c}{AUC}                                                  & \multicolumn{3}{c}{Accuracy}                                                  \\ \cline{2-7} 
\multicolumn{1}{c}{}                       & AS2015                 & NIPS34                 & POJ                    & AS2015                 & NIPS34                 & POJ                    \\ \hline
\textbf{DKT}                               & 0.7271±0.0005          & 0.7689±0.0002          & 0.6089±0.0009          & 0.7503±0.0003          & 0.7032±0.0004          & 0.6328±0.0020          \\
\textbf{DKT+}                              & \textbf{0.7285±0.0006}          & 0.7696±0.0002          & 0.6173±0.0007          & \underline{0.7510±0.0004}          & 0.7039±0.0004          & 0.6482±0.0021          \\
\textbf{KQN}                               & 0.7254±0.0004          & 0.7684±0.0003          & 0.6080±0.0015          & 0.7500±0.0003          & 0.7028±0.0001          & 0.6435±0.0017          \\
\textbf{DKVMN}                             & 0.7227±0.0004          & 0.7673±0.0004          & 0.6056±0.0022          & 0.7508±0.0006          & 0.7016±0.0005          & 0.6393±0.0015          \\
\textbf{ATKT}                              & 0.7245±0.0007          & 0.7665±0.0001          & 0.6075±0.0012          & 0.7494±0.0002          & 0.7013±0.0002          & 0.6332±0.0023          \\
\textbf{GKT}                               & 0.7258±0.0012          & 0.7689±0.0024          & 0.6070±0.0036          & 0.7504±0.0010          & 0.7014±0.0028          & 0.6117±0.0147          \\
\textbf{SAKT}                              & 0.7114±0.0003          & 0.7517±0.0005          & 0.6095±0.0013          & 0.7474±0.0002          & 0.6879±0.0004          & 0.6407±0.0035          \\
\textbf{SAINT}                             & 0.7026±0.0011          & 0.7873±0.0007          & 0.5563±0.0012          & 0.7438±0.0010          & 0.7180±0.0006          & 0.6476±0.0003          \\
\textbf{AKT}                               & \underline{0.7281±0.0004}          & \underline{0.8033±0.0003}          & \textbf{0.6281±0.0013}          & \textbf{0.7521±0.0005}          & \underline{0.7323±0.0005}          & 0.6492±0.0010          \\
\textbf{HAWKES}                            & -                      & 0.7767±0.0010          & -                      & -                      & 0.7110±0.0007          & -                      \\
\textbf{IEKT}                              & -                      & \textbf{0.8045±0.0002} & -                      & -                      & \textbf{0.7330±0.0002}   & -                      \\
\hline
\textbf{\textsc{sparseKT}-soft}                     & 0.7219±0.0006    & 0.7994±0.0005          & \underline{0.6247±0.0009}          & 0.7499±0.0004    & 0.7291±0.0005          &  \textbf{0.6512±0.0021}    \\
\textbf{\textsc{sparseKT}-topK}                     & 0.7203±0.0005 & 0.7997±0.0005    & 0.6218±0.0021 & 0.7499±0.0004 & 0.7287±0.0003      & \underline{0.6505±0.0037} \\
\bottomrule
\vspace{-0.6cm}
\end{tabular}
\label{tab:overall} 
\end{table}

\subsubsection{Impact of the Sparsity Level}



\begin{figure}[!tbph]
\centering
\vspace{-0.3cm}{}
\includegraphics[width=\linewidth]{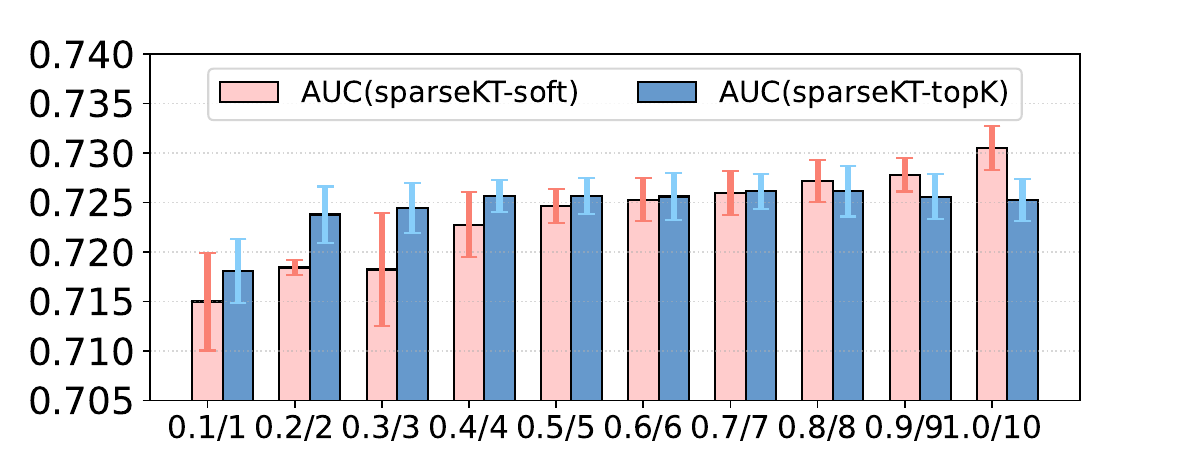}  \vspace{-0.6cm}
\caption{AUC performance of different values of $k$ with our \textsc{sparseKT}-soft and \textsc{sparseKT}-topK on AS2015.}
\label{fig:ablation}
\vspace{-0.5cm}
\end{figure}

We further explore the impacts of the sparse selection $k$ on the model performance. We conduct experiments on the AS2015 to evaluate our two sparse attention approaches with different values of $k$ on the validation set. The results are illustrated in Figure \ref{fig:ablation}. We limit the range of $k$ to $[0.1,1]$ and $[1,10]$ in soft-threshold and top-$K$ sparse attention respectively. We observe that neither soft-threshold nor top-$K$ sparse attention, the AUC scores are quite low when the values of $k$ are small, e.g., $k$=0.1/1 in soft-threshold and top-$K$ sparse attention respectively. We suppose that too limited historical interactions are selected and that the KT model can not obtain enough past learning information to predict students' future performance and hence get low AUC scores. With the increasing values of $k$, \textsc{sparseKT}-topK obtains better performance gradually and performs best AUC scores when $k=7$. After that, larger values of $k$ may contain more noise that decreases the model's robustness and limits its generalization yield to get a decreasing AUC score.



\subsubsection{Visualization of KC Relations via $k$-Sparse Attention}
\begin{figure}[!bpht]
\centering
\vspace{-0.2cm}
\includegraphics[width=\linewidth]{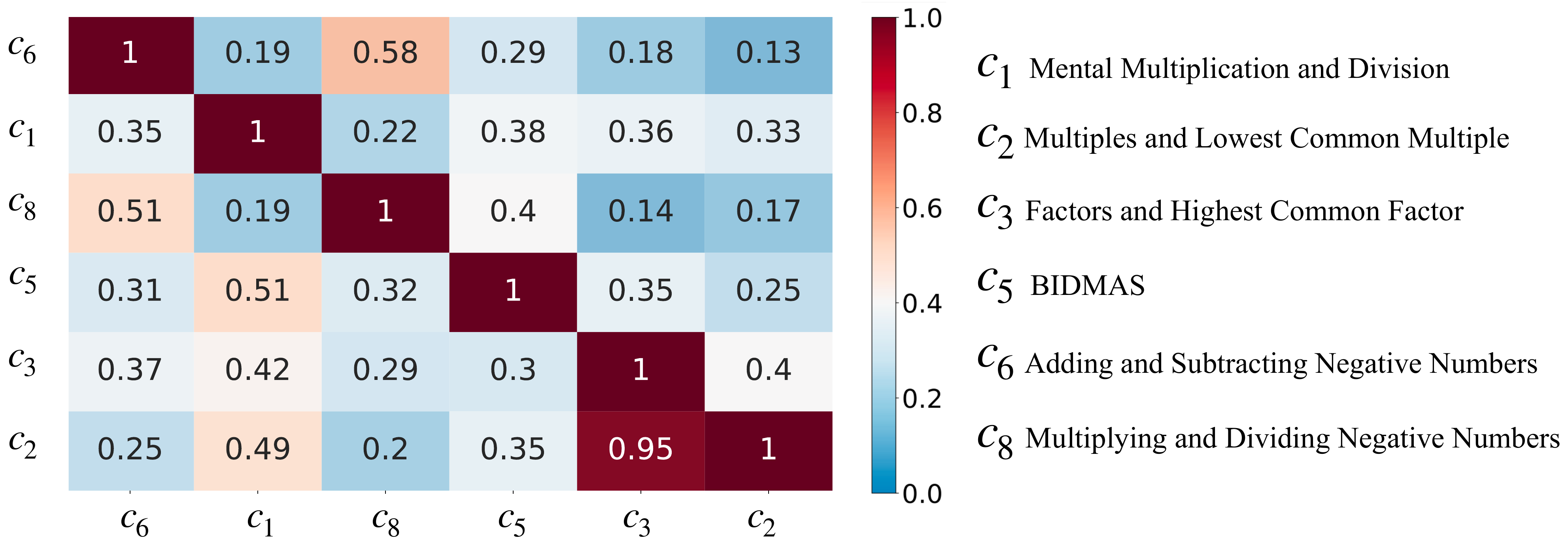}  \vspace{-0.6cm}
\caption{Attention weights visualization of \textsc{sparseKT}-topK. The y-axis is the pre-interaction KCs, and the x-axis is the post-interaction KCs.} \vspace{-0.4cm}
\label{fig:attn}
\end{figure}

Figure \ref{fig:attn} shows the KC relation visualization via our proposed $k$-sparse attention. We compute the cumulative sum of attention weights among all the KCs during the training of our \textsc{sparseKT}-soft. To better observe the relations, we compute the min-max normalization of the cumulative sum results. Due to the space limitation, we visualize the results between the top-6 KCs with the highest frequency on NIPS34. We can see that, since pre-post sequence relations among KCs, the attention weights are different in the same KC pairs. For example, for a KC pair <$c_2$,$c_3$>, $c_2$ has a high influence (0.95 weight) on $c3$ when $c_2$ is the pre-interaction of $c3$. On the contrary, $c_3$ has a relatively small impact (0.4 weight) on $c2$ when $c_3$ is the pre-interaction of $c_2$. Furthermore, there is a limited correlation to <$c_2$, $c_8$>, so the attention weights between them are less than 0.2 regardless of which KC is the pre-interaction.


\section{Conclusion}
\label{sec:conclusion}
In this paper, we propose \textsc{sparseKT} which enhances the classical scaled dot-product attention by extracting influential historical interactions to estimate students' mastery of knowledge. Extensive experimental results on three publicly educational datasets show the effectiveness and comparable prediction outcomes and robustness of \textsc{sparseKT}. In the future, we would like to explore more sparse attention approaches like dynamic $k$ selection or self-adaptive select $k$ attention weights without the hyperparameter tuning.

\begin{acks}
This work was supported in part by National Key R\&D Program of China, under Grant No. 2020AAA0104500; in part by Beijing Nova Program (Z201100006820068) from Beijing Municipal Science \& Technology Commission; in part by NFSC under Grant No. 61877029; in part by Key Laboratory of Smart Education of Guangdong Higher Education Institutes, Jinan University (2022LSY-S003) and in part by National Joint Engineering Research Center of Network Security Detection and Protection Technology.
\end{acks}

\bibliographystyle{ACM-Reference-Format}
\balance
\bibliography{sigir2023}


\begin{thebibliography}{24}


\ifx \showCODEN    \undefined \def \showCODEN     #1{\unskip}     \fi
\ifx \showDOI      \undefined \def \showDOI       #1{#1}\fi
\ifx \showISBNx    \undefined \def \showISBNx     #1{\unskip}     \fi
\ifx \showISBNxiii \undefined \def \showISBNxiii  #1{\unskip}     \fi
\ifx \showISSN     \undefined \def \showISSN      #1{\unskip}     \fi
\ifx \showLCCN     \undefined \def \showLCCN      #1{\unskip}     \fi
\ifx \shownote     \undefined \def \shownote      #1{#1}          \fi
\ifx \showarticletitle \undefined \def \showarticletitle #1{#1}   \fi
\ifx \showURL      \undefined \def \showURL       {\relax}        \fi
\providecommand\bibfield[2]{#2}
\providecommand\bibinfo[2]{#2}
\providecommand\natexlab[1]{#1}
\providecommand\showeprint[2][]{arXiv:#2}

\bibitem[Abdelrahman and Wang(2019)]%
        {abdelrahman2019knowledge}
\bibfield{author}{\bibinfo{person}{Ghodai Abdelrahman} {and}
  \bibinfo{person}{Qing Wang}.} \bibinfo{year}{2019}\natexlab{}.
\newblock \showarticletitle{Knowledge tracing with sequential key-value memory
  networks}. In \bibinfo{booktitle}{\emph{Proceedings of the 42nd International
  ACM SIGIR Conference on Research and Development in Information Retrieval}}.
  \bibinfo{pages}{175--184}.
\newblock


\bibitem[Amini and Ghaemmaghami(2022)]%
        {amini2022towards}
\bibfield{author}{\bibinfo{person}{Sajjad Amini} {and}
  \bibinfo{person}{Shahrokh Ghaemmaghami}.} \bibinfo{year}{2022}\natexlab{}.
\newblock \showarticletitle{Towards Robust Visual Transformer Networks via
  K-Sparse Attention}. In \bibinfo{booktitle}{\emph{ICASSP 2022-2022 IEEE
  International Conference on Acoustics, Speech and Signal Processing}}. IEEE,
  \bibinfo{pages}{4053--4057}.
\newblock


\bibitem[Chen et~al\mbox{.}(2023)]%
        {chen2023improving}
\bibfield{author}{\bibinfo{person}{Jiahao Chen}, \bibinfo{person}{Zitao Liu},
  \bibinfo{person}{Shuyan Huang}, \bibinfo{person}{Qiongqiong Liu}, {and}
  \bibinfo{person}{Weiqi Luo}.} \bibinfo{year}{2023}\natexlab{}.
\newblock \showarticletitle{Improving Interpretability of Deep Sequential
  Knowledge Tracing Models with Question-centric Cognitive Representations}. In
  \bibinfo{booktitle}{\emph{Proceedings of the AAAI Conference on Artificial
  Intelligence}}.
\newblock


\bibitem[Child et~al\mbox{.}(2019)]%
        {child2019generating}
\bibfield{author}{\bibinfo{person}{Rewon Child}, \bibinfo{person}{Scott Gray},
  \bibinfo{person}{Alec Radford}, {and} \bibinfo{person}{Ilya Sutskever}.}
  \bibinfo{year}{2019}\natexlab{}.
\newblock \showarticletitle{Generating long sequences with sparse
  transformers}.
\newblock \bibinfo{journal}{\emph{arXiv preprint arXiv:1904.10509}}
  (\bibinfo{year}{2019}).
\newblock


\bibitem[Choi et~al\mbox{.}(2020)]%
        {choi2020towards}
\bibfield{author}{\bibinfo{person}{Youngduck Choi}, \bibinfo{person}{Youngnam
  Lee}, \bibinfo{person}{Junghyun Cho}, \bibinfo{person}{Jineon Baek},
  \bibinfo{person}{Byungsoo Kim}, \bibinfo{person}{Yeongmin Cha},
  \bibinfo{person}{Dongmin Shin}, \bibinfo{person}{Chan Bae}, {and}
  \bibinfo{person}{Jaewe Heo}.} \bibinfo{year}{2020}\natexlab{}.
\newblock \showarticletitle{Towards an appropriate query, key, and value
  computation for knowledge tracing}. In \bibinfo{booktitle}{\emph{Proceedings
  of the Seventh ACM Conference on Learning@ Scale}}.
  \bibinfo{pages}{341--344}.
\newblock


\bibitem[Ghosh et~al\mbox{.}(2020)]%
        {ghosh2020context}
\bibfield{author}{\bibinfo{person}{Aritra Ghosh}, \bibinfo{person}{Neil
  Heffernan}, {and} \bibinfo{person}{Andrew~S Lan}.}
  \bibinfo{year}{2020}\natexlab{}.
\newblock \showarticletitle{Context-aware attentive knowledge tracing}. In
  \bibinfo{booktitle}{\emph{Proceedings of the 26th ACM SIGKDD International
  Conference on Knowledge Discovery \& Data Mining}}.
  \bibinfo{pages}{2330--2339}.
\newblock


\bibitem[Guo et~al\mbox{.}(2021)]%
        {guo2021enhancing}
\bibfield{author}{\bibinfo{person}{Xiaopeng Guo}, \bibinfo{person}{Zhijie
  Huang}, \bibinfo{person}{Jie Gao}, \bibinfo{person}{Mingyu Shang},
  \bibinfo{person}{Maojing Shu}, {and} \bibinfo{person}{Jun Sun}.}
  \bibinfo{year}{2021}\natexlab{}.
\newblock \showarticletitle{Enhancing Knowledge Tracing via Adversarial
  Training}. In \bibinfo{booktitle}{\emph{Proceedings of the 29th ACM
  International Conference on Multimedia}}. \bibinfo{pages}{367--375}.
\newblock


\bibitem[Lee and Yeung(2019)]%
        {lee2019knowledge}
\bibfield{author}{\bibinfo{person}{Jinseok Lee} {and} \bibinfo{person}{Dit-Yan
  Yeung}.} \bibinfo{year}{2019}\natexlab{}.
\newblock \showarticletitle{Knowledge query network for knowledge tracing: How
  knowledge interacts with skills}. In \bibinfo{booktitle}{\emph{Proceedings of
  the 9th International Conference on Learning Analytics \& Knowledge}}.
  \bibinfo{pages}{491--500}.
\newblock


\bibitem[Liu et~al\mbox{.}(2023b)]%
        {liu2023enhancing}
\bibfield{author}{\bibinfo{person}{Zitao Liu}, \bibinfo{person}{Qiongqiong
  Liu}, \bibinfo{person}{Jiahao Chen}, \bibinfo{person}{Shuyan Huang},
  \bibinfo{person}{Boyu Gao}, \bibinfo{person}{Weiqi Luo}, {and}
  \bibinfo{person}{Jian Weng}.} \bibinfo{year}{2023}\natexlab{b}.
\newblock \showarticletitle{Enhancing Deep Knowledge Tracing with Auxiliary
  Tasks}. In \bibinfo{booktitle}{\emph{Proceedings of the ACM Web Conference
  2023}}.
\newblock


\bibitem[Liu et~al\mbox{.}(2023a)]%
        {liusimplekt}
\bibfield{author}{\bibinfo{person}{Zitao Liu}, \bibinfo{person}{Qiongqiong
  Liu}, \bibinfo{person}{Jiahao Chen}, \bibinfo{person}{Shuyan Huang}, {and}
  \bibinfo{person}{Weiqi Luo}.} \bibinfo{year}{2023}\natexlab{a}.
\newblock \showarticletitle{simpleKT: A Simple But Tough-to-Beat Baseline for
  Knowledge Tracing}. In \bibinfo{booktitle}{\emph{The Eleventh International
  Conference on Learning Representations}}.
\newblock


\bibitem[Liu et~al\mbox{.}(2022)]%
        {liu2022pykt}
\bibfield{author}{\bibinfo{person}{Zitao Liu}, \bibinfo{person}{Qiongqiong
  Liu}, \bibinfo{person}{Jiahao Chen}, \bibinfo{person}{Shuyan Huang},
  \bibinfo{person}{Jiliang Tang}, {and} \bibinfo{person}{Weiqi Luo}.}
  \bibinfo{year}{2022}\natexlab{}.
\newblock \showarticletitle{py{KT}: A Python Library to Benchmark Deep Learning
  based Knowledge Tracing Models}. In \bibinfo{booktitle}{\emph{Thirty-sixth
  Conference on Neural Information Processing Systems Datasets and Benchmarks
  Track}}.
\newblock


\bibitem[Long et~al\mbox{.}(2021)]%
        {long2021tracing}
\bibfield{author}{\bibinfo{person}{Ting Long}, \bibinfo{person}{Yunfei Liu},
  \bibinfo{person}{Jian Shen}, \bibinfo{person}{Weinan Zhang}, {and}
  \bibinfo{person}{Yong Yu}.} \bibinfo{year}{2021}\natexlab{}.
\newblock \showarticletitle{Tracing Knowledge State with Individual Cognition
  and Acquisition Estimation}. In \bibinfo{booktitle}{\emph{Proceedings of the
  44th International ACM SIGIR Conference on Research and Development in
  Information Retrieval}}. \bibinfo{pages}{173--182}.
\newblock


\bibitem[Martins and Astudillo(2016)]%
        {martins2016softmax}
\bibfield{author}{\bibinfo{person}{Andre Martins} {and} \bibinfo{person}{Ramon
  Astudillo}.} \bibinfo{year}{2016}\natexlab{}.
\newblock \showarticletitle{From softmax to sparsemax: A sparse model of
  attention and multi-label classification}. In
  \bibinfo{booktitle}{\emph{International Conference on Machine Learning}}.
  PMLR, \bibinfo{pages}{1614--1623}.
\newblock


\bibitem[Nakagawa et~al\mbox{.}(2019)]%
        {nakagawa2019graph}
\bibfield{author}{\bibinfo{person}{Hiromi Nakagawa}, \bibinfo{person}{Yusuke
  Iwasawa}, {and} \bibinfo{person}{Yutaka Matsuo}.}
  \bibinfo{year}{2019}\natexlab{}.
\newblock \showarticletitle{Graph-based knowledge tracing: modeling student
  proficiency using graph neural network}. In \bibinfo{booktitle}{\emph{2019
  IEEE/WIC/ACM International Conference on Web Intelligence}}. IEEE,
  \bibinfo{pages}{156--163}.
\newblock


\bibitem[Pandey and Karypis(2019)]%
        {pandey2019self}
\bibfield{author}{\bibinfo{person}{Shalini Pandey} {and}
  \bibinfo{person}{George Karypis}.} \bibinfo{year}{2019}\natexlab{}.
\newblock \showarticletitle{A self-attentive model for knowledge tracing}. In
  \bibinfo{booktitle}{\emph{12th International Conference on Educational Data
  Mining}}. International Educational Data Mining Society,
  \bibinfo{pages}{384--389}.
\newblock


\bibitem[Pandey and Srivastava(2020)]%
        {pandey2020rkt}
\bibfield{author}{\bibinfo{person}{Shalini Pandey} {and}
  \bibinfo{person}{Jaideep Srivastava}.} \bibinfo{year}{2020}\natexlab{}.
\newblock \showarticletitle{{RKT}: relation-aware self-attention for knowledge
  tracing}. In \bibinfo{booktitle}{\emph{Proceedings of the 29th ACM
  International Conference on Information \& Knowledge Management}}.
  \bibinfo{pages}{1205--1214}.
\newblock


\bibitem[Peters et~al\mbox{.}(2019)]%
        {peters2019sparse}
\bibfield{author}{\bibinfo{person}{Ben Peters}, \bibinfo{person}{Vlad Niculae},
  {and} \bibinfo{person}{Andr{\'e}~FT Martins}.}
  \bibinfo{year}{2019}\natexlab{}.
\newblock \showarticletitle{Sparse Sequence-to-Sequence Models}. In
  \bibinfo{booktitle}{\emph{Proceedings of the 57th Annual Meeting of the
  Association for Computational Linguistics}}. \bibinfo{pages}{1504--1519}.
\newblock


\bibitem[Piech et~al\mbox{.}(2015)]%
        {piech2015deep}
\bibfield{author}{\bibinfo{person}{Chris Piech}, \bibinfo{person}{Jonathan
  Bassen}, \bibinfo{person}{Jonathan Huang}, \bibinfo{person}{Surya Ganguli},
  \bibinfo{person}{Mehran Sahami}, \bibinfo{person}{Leonidas~J Guibas}, {and}
  \bibinfo{person}{Jascha Sohl-Dickstein}.} \bibinfo{year}{2015}\natexlab{}.
\newblock \showarticletitle{Deep knowledge tracing}.
\newblock \bibinfo{journal}{\emph{Advances in Neural Information Processing
  Systems}}  \bibinfo{volume}{28} (\bibinfo{year}{2015}).
\newblock


\bibitem[Vaswani et~al\mbox{.}(2017)]%
        {vaswani2017attention}
\bibfield{author}{\bibinfo{person}{Ashish Vaswani}, \bibinfo{person}{Noam
  Shazeer}, \bibinfo{person}{Niki Parmar}, \bibinfo{person}{Jakob Uszkoreit},
  \bibinfo{person}{Llion Jones}, \bibinfo{person}{Aidan~N Gomez},
  \bibinfo{person}{{\L}ukasz Kaiser}, {and} \bibinfo{person}{Illia
  Polosukhin}.} \bibinfo{year}{2017}\natexlab{}.
\newblock \showarticletitle{Attention is all you need}.
\newblock \bibinfo{journal}{\emph{Advances in Neural Information Processing
  Systems}}  \bibinfo{volume}{30} (\bibinfo{year}{2017}).
\newblock


\bibitem[Wang et~al\mbox{.}(2021)]%
        {wang2021temporal}
\bibfield{author}{\bibinfo{person}{Chenyang Wang}, \bibinfo{person}{Weizhi Ma},
  \bibinfo{person}{Min Zhang}, \bibinfo{person}{Chuancheng Lv},
  \bibinfo{person}{Fengyuan Wan}, \bibinfo{person}{Huijie Lin},
  \bibinfo{person}{Taoran Tang}, \bibinfo{person}{Yiqun Liu}, {and}
  \bibinfo{person}{Shaoping Ma}.} \bibinfo{year}{2021}\natexlab{}.
\newblock \showarticletitle{Temporal cross-effects in knowledge tracing}. In
  \bibinfo{booktitle}{\emph{Proceedings of the 14th ACM International
  Conference on Web Search and Data Mining}}. \bibinfo{pages}{517--525}.
\newblock


\bibitem[Wang et~al\mbox{.}(2020)]%
        {wang2020instructions}
\bibfield{author}{\bibinfo{person}{Zichao Wang}, \bibinfo{person}{Angus Lamb},
  \bibinfo{person}{Evgeny Saveliev}, \bibinfo{person}{Pashmina Cameron},
  \bibinfo{person}{Yordan Zaykov}, \bibinfo{person}{Jos{\'e}~Miguel
  Hern{\'a}ndez-Lobato}, \bibinfo{person}{Richard~E Turner},
  \bibinfo{person}{Richard~G Baraniuk}, \bibinfo{person}{Craig Barton},
  \bibinfo{person}{Simon~Peyton Jones}, {et~al\mbox{.}}}
  \bibinfo{year}{2020}\natexlab{}.
\newblock \showarticletitle{Instructions and Guide for Diagnostic Questions:
  The NeurIPS 2020 Education Challenge}.
\newblock \bibinfo{journal}{\emph{ArXiv preprint}}
  \bibinfo{volume}{abs/2007.12061} (\bibinfo{year}{2020}).
\newblock
\urldef\tempurl%
\url{https://arxiv.org/abs/2007.12061}
\showURL{%
\tempurl}


\bibitem[Yeung and Yeung(2018)]%
        {yeung2018addressing}
\bibfield{author}{\bibinfo{person}{Chun-Kit Yeung} {and}
  \bibinfo{person}{Dit-Yan Yeung}.} \bibinfo{year}{2018}\natexlab{}.
\newblock \showarticletitle{Addressing two problems in deep knowledge tracing
  via prediction-consistent regularization}. In
  \bibinfo{booktitle}{\emph{Proceedings of the Fifth Annual ACM Conference on
  Learning at Scale}}. \bibinfo{pages}{1--10}.
\newblock


\bibitem[Zhang et~al\mbox{.}(2017)]%
        {zhang2017dynamic}
\bibfield{author}{\bibinfo{person}{Jiani Zhang}, \bibinfo{person}{Xingjian
  Shi}, \bibinfo{person}{Irwin King}, {and} \bibinfo{person}{Dit-Yan Yeung}.}
  \bibinfo{year}{2017}\natexlab{}.
\newblock \showarticletitle{Dynamic key-value memory networks for knowledge
  tracing}. In \bibinfo{booktitle}{\emph{Proceedings of the 26th International
  Conference on World Wide Web}}. \bibinfo{pages}{765--774}.
\newblock


\bibitem[Zhao et~al\mbox{.}(2019)]%
        {zhao2019explicit}
\bibfield{author}{\bibinfo{person}{Guangxiang Zhao}, \bibinfo{person}{Junyang
  Lin}, \bibinfo{person}{Zhiyuan Zhang}, \bibinfo{person}{Xuancheng Ren},
  \bibinfo{person}{Qi Su}, {and} \bibinfo{person}{Xu Sun}.}
  \bibinfo{year}{2019}\natexlab{}.
\newblock \showarticletitle{Explicit sparse transformer: Concentrated attention
  through explicit selection}.
\newblock \bibinfo{journal}{\emph{arXiv preprint arXiv:1912.11637}}
  (\bibinfo{year}{2019}).
\newblock


\end{thebibliography}

\end{document}